\documentclass[journal]{IEEEtran}
\usepackage{caption}
\usepackage{amsmath,amssymb,amsfonts}
\usepackage{algorithmic}
\usepackage{graphicx}
\usepackage{textcomp}
\usepackage{xcolor}
\usepackage{diagbox} 
\usepackage{verbatim}
\usepackage{comment}
\usepackage[utf8]{inputenc}
\usepackage[english]{babel}
\usepackage{float}
\usepackage[font=footnotesize]{subfig}
\usepackage{verbatim}
\usepackage{float}
\usepackage{mathtools}
\DeclarePairedDelimiter{\norm}{\lVert}{\rVert}
\ifCLASSINFOpdf
\else
\fi
\begin{document}
\title{CoFF: Cooperative Spatial Feature Fusion for 3D Object Detection on Autonomous Vehicles}

\author{Jingda Guo, Dominic Carrillo, Sihai Tang, Qi Chen, 
Qing Yang,~\IEEEmembership{Senior Member, IEEE,}
Song Fu,~\IEEEmembership{Senior Member,~IEEE,}
Xi Wang~\IEEEmembership{Member,~IEEE,}
Nannan Wang~\IEEEmembership{Member,~IEEE,}
Paparao Palacharla~\IEEEmembership{Member,~IEEE} \vspace{-25pt}
\thanks{J. Guo, D. Carrillo, S. Tang, Q. Chen, Q. Yang, and S. Fu are with the Department of Computer Science and Engineering, University of North Texas, Deton, TX, 76207 USA. e-mail: (JingdaGuo@my.unt.edu, DominicCarrillo@my.unt.edu, qing.yang@unt.edu, song.fu@unt.edu).}% <-this % stops a space
\thanks{X. Wang, N. Wang and P. Palacharla are with Fujitsu.}% <-this % stops a space
\thanks{Manuscript received April 19, 20xx; revised August 26, 20xx.}}

% a space would be appended to the last name and could cause e name on that

\markboth{Journal of \LaTeX\ Class Files,~Vol.~14, No.~8, August~2015}%
{Shell \MakeLowercase{\textit{et al.}}: CoFF: Cooperative Spatial Feature Fusion for 3D Object Detection on Autonomous Vehicles}



\maketitle

\begin{abstract}
%Cooperative perception provides a novel way to improve the driving safety on autonomous vehicles (AV) by considering information shared from other nearby AVs. 
To reduce the amount of transmitted data, feature map based fusion is recently proposed as a practical solution to cooperative 3D object detection by autonomous vehicles. The precision of object detection, however, may require significant improvement, especially for objects that are far away or occluded. To address this critical issue for the safety of autonomous vehicles and human beings, we propose a cooperative spatial feature fusion (CoFF) method for autonomous vehicles to effectively fuse feature maps for achieving a higher 3D object detection performance. Specially, CoFF differentiates weights among feature maps for a more guided fusion, based on how much new semantic information is provided by the received feature maps. It also enhances the inconspicuous features corresponding to far/occluded objects to improve their detection precision. Experimental results show that CoFF achieves a significant improvement in terms of both detection precision and effective detection range for autonomous vehicles, compared to previous feature fusion solutions.

\end{abstract}


% \fi
\IEEEpeerreviewmaketitle

\vspace{-11pt}
\section{Introduction}
%Interacting with environments and other intelligent facilities, an autonomous vehicle (AV) can free human drivers from complex driving tasks, and reduce the chance of accidents. 
%
An autonomous vehicle (AV) relies on its perception system to sense the surroundings and makes driving decisions accordingly.
However, the sensors equipped on AVs are typically non-line-of-sight, whose effective sensing range could be significantly reduced on crowded roads, due to blockages and/or occlusions. 
Therefore, it is crucial to connect AVs and allow them to exchange sensor data to facilitate precise cooperative perception, thus improving driving safety.
A major challenge for cooperative perception on connected and autonomous vehicles (CAVs) is how to effectively merge sensor data received from different AVs to obtain a precise and comprehensive perception.

Due to the huge volume of raw sensor data, it is practically infeasible to exchange raw data among vehicles, which would cause severe bottlenecks in existing network infrastructures.
To reduce network traffic, a feature map based data sharing mechanism is proposed for 3D object detection on autonomous vehicles~\cite{chen2019f}.
Feature maps are the intermediate results produced by a Convolutional Neural Network (CNN).
In~\cite{chen2019f}, feature maps generated on different vehicles are combined to yield a cooperative object detection. However, we find that its feature fusion mechanism can be further improved, by considering the volume of new semantic information contained in the to-be-fused feature maps. 
In this article, we aim to achieve an enhanced object detection performance by fusing feature maps in an intelligent manner.
%In this article, we focus on 3D object detection in the perception system of CAVs.

Unlike raw sensor data, feature maps are hard to interpret, which increases the difficulty in designing an effective fusion mechanism for cooperative 3D object detection. 
To tackle this challenge, we investigate how the importance of a received feature map is influenced by the distance of the vehicle from which the feature map is generated. We call this approach 'cooperative spatial feature fusion.'
We hypothesize that feature maps produced by a distant vehicle can significantly improve the object detection on the current vehicle, particularly, for recognizing distant objects.
Moreover, a better object detection result is expected if noise signals in feature maps can be reduced or removed.
Towards this end, we propose a novel cooperative spatial feature fusion mechanism for CAVs to effectively fuse feature maps and achieve accurate 3D object detection.

\vspace{-11pt}
\subsection{Proposed Solution}

In designing the spatial feature fusion method, we need to conquer two major technical challenges.
The first challenge is how to identify and reduce the negative effects on object detection caused by mistakenly fused feature maps generated by different vehicles.
This problem was overlooked in the existing work~\cite{chen2019f} as it does not consider how feature maps affect each other when they are fused in a wrong way.
The underlying fusing function adopted by the previous work is  $\textit{maxout}$ which selects the features with larger values in the fusion process.
The method seems reasonable as it keeps the most distinctive features while suppressing non-distinctive ones; however, it might omit important features received from other vehicles, which could have significantly improved the current vehicle's object detection performance if used correctly.
In other words, the feature maps generated by multiple vehicles should be treated differently, instead of equally as was the case in the previous work~\cite{chen2019f}.

The second challenge lies in the difficulty of detecting distant or occluded objects. 
This is a problem not just for cooperative object detection, but a common issue for many 3D object detection approaches for autonomous vehicles. 
When feature maps are fused and more information is considered, we can detect objects which are otherwise hard to detect due to insufficient information from individual sensor data.
This is because feature maps generated by different vehicles complement each other, and if fused properly, they can offer a more comprehensive representation of objects.
%Towards this end, we design a feature enhancement mechanism which makes important features (representing objects) more appealing while keeping the noise features hideous.
%

%By solving the above-mentioned challenges, we design an enhanced feature-map based fusion mechanisam (CoFF) in which constitute by two parts: information-based feature map fusion and feature enhancement.
%information-based feature fusion can reduce the effect from receivers' feature map, and keep important feature from both feature map after fusion. feature enhancement enhance the weak feature from previous undetected objects which may caused by distance/occlusion, and improve the precision on detecting those objects.
%

\begin{figure*}[!htbp]
\centering
\vspace{-3pt}{%
\includegraphics[width=0.75\textwidth]{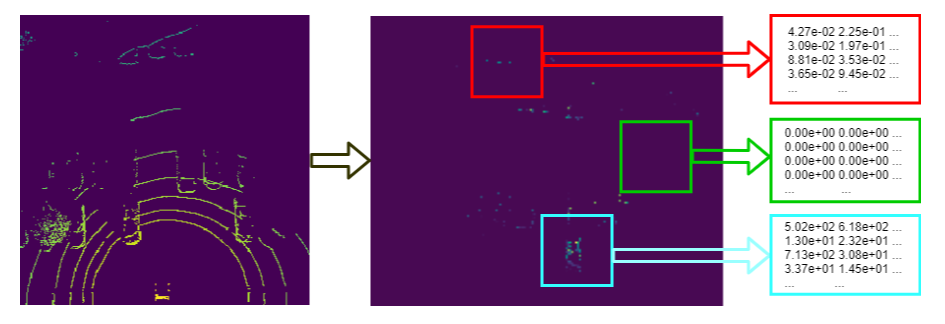}
}
\vspace{-8pt}
\caption{Illustration of a feature map generated by F-Cooper~\cite{chen2019f} from LiDAR point cloud data. Examples of strong features, weak features, and background features are depicted in blue, red, and green boxes, respectively.}
\vspace{-15pt}
\label{fig:value}
\end{figure*}

\vspace{-11pt}
\subsection{Contributions}
The main contributions of this article can be summarized as follows. 
First, we propose a new feature fusion approach to cooperative perception on autonomous vehicles, which aims to improve the 3D object detection performance for distant, or occluded objects.
Our novel idea is to factor in the new semantic information when fusing feature maps from different vehicles, e.g., greater weights are given to the feature maps that contains more new semantic information compared with its own feature map. 
Second, we discover that the numerical values in a feature map usually represent the significance of the underlying features on 3D detection.
Thus, enlarging the difference between the features representing the objects and those representing the background can help detect objects.
Our proposed feature enhancement method aims at increasing the values of the features representing objects, while keeping the background features almost unchanged.
Our proposed method is generic and applicable to other applications that involve fusing data/features generated by different sensors/entities, e.g.,  in Internet of Things environments.
%

%The paper is organized as follows. In section \uppercase\expandafter{\romannumeral2}, we discuss the state-of-the-art solution for cooperative perception on autonomous, and their limitation. In section \uppercase\expandafter{\romannumeral3}, we proposed our enhanced feature fusion approach. In section \uppercase\expandafter{\romannumeral4}, we introduce our dataset, test cases and results in experiment. We concluded CoFF and indicate future work in section \uppercase\expandafter{\romannumeral5}.

\vspace{-7pt}

\section{Preliminaries and Background}
It has been shown that sharing raw LiDAR (Light Detection and Ranging) data among autonomous vehicles can help 3D object detection on individual vehicles.
The basic idea of cooperative perception, Cooper~\cite{chen2019cooper}, is to fuse LiDAR point cloud data produced by multiple vehicles to cooperatively detect 3D objects.
While Cooper~\cite{chen2019cooper} provides a means for raw sensor data fusion to improve object detection performance, transmitting raw point cloud data places a heavy burden on vehicle-to-vehicle (V2V) wireless networks. 
One frame of 3D point cloud can be as large as 3 MB, and a typical LiDAR can generate as many as 20 frames per second, equivalent to 480 Mbps of network capacity. 
Therefore, it is difficult to continuously transmit such massive amount of data over today's wireless networks.
As an intermediate result from a CNN-based detection model, feature maps carry important semantic information for object detection and can be considered a substitute for the original sensor data for cooperative object detection. 

In this section, we briefly present the concept of feature maps in Section~\ref{sec:fm}, and then describe feature maps for 3D object detection in Section~\ref{sec:f2}. We explain feature map based fusion for cooperative perception on AVs in Section~\ref{sec:f3} and discuss the limitations of existing approaches in Section~\ref{sec:f4}.

\vspace{-5pt}
\subsection{Feature Maps from CNN}
\label{sec:fm}

With the rapid growth and increasing capability of CNN, most state-of-the-art object detection models, such as \cite{redmon2018yolov3,shi2019pointrcnn,ku2018joint}, on autonomous vehicles are CNN-based. 
A major component of CNN models is the feature extractor which is composed of multiple convolutional layers. 
The feature extractor is trained using vast amount of labeled data, and has the capability of extracting features from the original sensor data for multiple recognizing tasks.
A feature map is the output of the feature extractor~\cite{girshick2015fast}, and thus contains important semantic information to accomplish object recognizing tasks. 
Features in a feature map are uninterpretable. However, feature maps provide necessary semantic information for further processing, with only being a fraction of the original data size.

\vspace{-11pt}
\subsection{Feature Maps for 3D Object Detection}
\label{sec:f2}

For 3D object detection, many pioneers' works \cite{engelcke2017vote3deep,li20173d,zhou2018voxelnet} divide 3D space into voxels and generate corresponding location-aware features using 3D convolutions. 
As a representative solution to CNN-based 3D object detection, VoxelNet~\cite{zhou2018voxelnet} becomes the backbone network for many state-of-the-art 3D detectors~(e.g., \cite{yan2018second,lang2019pointpillars}). 
The feature extractor of F-Cooper \cite{chen2019f} leverages the design of VoxelNet. 
%Therefore, we focus in this article on the feature maps produced from the F-Cooper.
%
F-Cooper takes the original 3D point cloud data as input and divides it into thousands of voxels, each of which contains a subset of point clouds in a three-dimensional space.
Then, F-Cooper processes these voxels to extract features and build a feature map which is used for object detection.

Fig.~\ref{fig:value} shows an example of a spatial feature map generated from VoxelNet on LiDAR data. 
Voxels containing more point clouds typically show prominent features, e.g., those depicted in the blue box in the figure have greater values on average than those in the feature map.
In contrast, the values enclosed in the red box are smaller than those in the blue box, indicating fewer point cloud data collected from that region.  
For the voxels containing no point cloud data, i.e., they do not provide any useful information for object detection, their corresponding values in the feature map are all zeros, as shown in the green box.
We define a feature in the feature map that contains more larger values as a \textbf{strong feature}, while a feature with fewer larger values as a \textbf{weak feature}.
We will use these definitions to compare the importance of two feature maps, which will be crucial for designing the feature fusion mechanism in later discussion.

\begin{figure*}[!htbp]
\centering
\vspace{-11pt}
\subfloat[3D data of receiver]{%
\includegraphics[width=35mm,height=60mm]{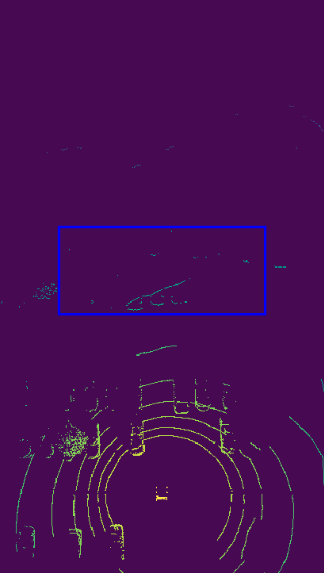}
}
\subfloat[3D data of sender]{%
\includegraphics[width=35mm,height=60mm]{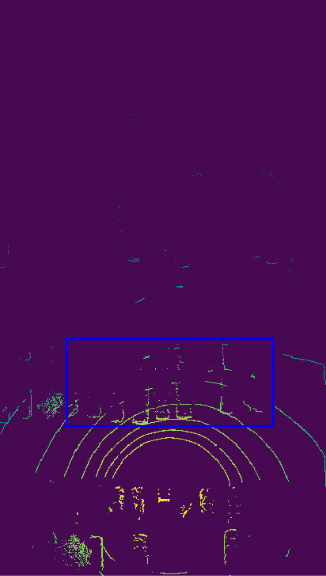}
}
\subfloat[Detection results on receiver]{%
\includegraphics[width=35mm,height=60mm]{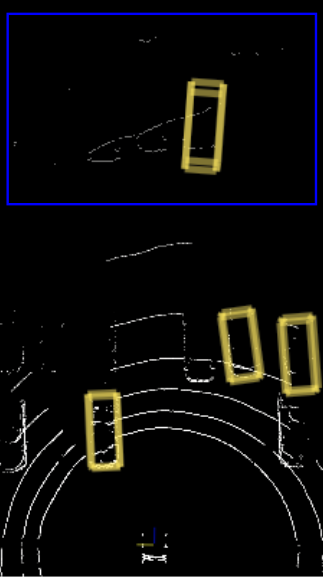}
}
\subfloat[Detection results on sender]{%
\includegraphics[width=35mm,height=60mm]{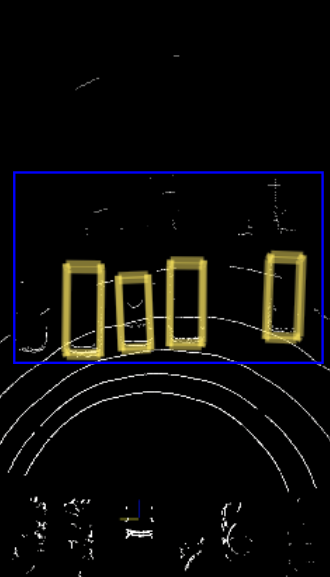}
}
\subfloat[F-Cooper's results]{%
\includegraphics[width=35mm,height=60mm]{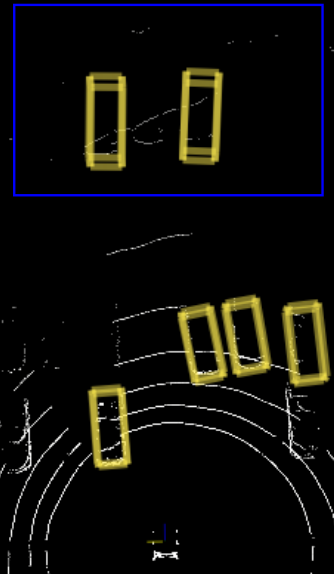}
}
\vspace{-3pt}
\caption{Limitations of F-Cooper fusion on detecting far objects. Detection results on far objects of F-Cooper is not as good as those on the sender side, shown in the blue box. }
\vspace{-11pt}
\label{fig:limit_f}
\end{figure*}

\vspace{-11pt}
\subsection{Feature Map Based Fusion}
\label{sec:f3}
A feature map can be considered a substitute for the original sensor data; therefore, cooperative perception on CAVs can also be achieved by fusing feature maps. 
F-Cooper~\cite{chen2019f} is a state-of-the-art solution that achieves cooperative 3D object detection by sharing feature maps among autonomous vehicles. 
For an AV that receives a feature map from a nearby AV, it fuses the received feature map with its own feature map by aligning them based on their physical locations.
%For a certain physical region, as long as sender and receiver are using same detection model/feature extractor, the feature on two feature maps is similar. 
%Therefore, feature map fusion provide an opportunity to fuse the similar feature together and improve the detection performance.
%
The location information can be obtained from the corresponding point cloud data.
The fusion of the feature maps can be viewed as creating a new feature map that contains merged features.
Experimental results show that the $maxout$ function~\cite{goodfellow2013maxout} helps F-Cooper detect more objects from the fused feature maps, including objects that cannot be detected by individual sender/receiver AV. 

\vspace{-11pt}
\subsection{Limitations of F-Cooper}
\label{sec:f4}

There are two major limitations originated from the $maxout$ fusion function in F-Cooper~\cite{chen2019f}.
First, the $maxout$ function does not consider the importance of individual to-be-fused feature maps.
Second, F-Cooper tends to have difficulty in detecting distant or occluded objects. 

A illustrative example is shown in Fig.~\ref{fig:limit_f}, in which a sender vehicle is sharing its sensor data, in the feature map format, to a receiver vehicle located behind the sender vehicle.
For the same region (depicted in the blue box in Fig.~\ref{fig:limit_f}(a) and (b)), stronger features are more likely to be generated by the sender, as shown in Fig.~\ref{fig:limit_f}(b), due to its physical proximity to the region.
This suggests that the sender would have a better object detection performance for this region, as shown in Fig.~\ref{fig:limit_f}(c) and (d).
For the same region, as it is relatively far from the receiver, as shown in Fig.~\ref{fig:limit_f}(a), the receiver would generate weak features on its feature map. 
Due to laser scattering and occlusion, however, some values for that region in the receiver's feature map could be larger than those in the sender's feature map.
As the $maxout$ function essentially keeps larger values, certain parts of the weak features from receiver's feature map would be kept, while the corresponding features provided by the sender will be removed.
As the \textit{maxout} function treats all feature maps equally, weak features from the receiver's feature map will affect the overall detection performance on the fused feature map. 
As shown in Fig.~\ref{fig:limit_f}(e), two objects that were detected by the sender do not show up in the detection results after fusion.   
%Fig.~\ref{fig:limit_f} shows an example of this negative effect of \textit{maxout}. Blue box marks the same area from both sender and receiver, and sender has a good detection results on this area as shown on Fig.~\ref{fig:limit_f}(b). Due to the distance and occlusion, the feature on receiver's feature map is weak and affect the F-Cooper fusion results. The detection results for that area is not as good as sender's, as shown on Fig.~\ref{fig:limit_f}(e). 
%
The negative effects of weak features on cooperative object detection become more prominent when occlusions occur more frequently, e.g., in a heavy traffic scenario.
%The precision of F-Cooper is impervious when sender and receiver is close, but the detection performance decay rapidly along with the distance increase. 
%Thus, simply using \textit{maxout} for all fusion cases without considering the correlation of feature maps may not be a good way. 

%Close objects reflect more points to LiDAR, which generate prominent feature on feature map, and makes them easier to be detected. The detection performance is fairly good for close objects, even without cooperative perception. 
%
As laser signals are more likely to be blocked by nearby objects or effuse into the environment, far/occluded objects generally reflect few laser signals back, thus generating less point cloud data. 
In other words, these objects are represented as relatively inconspicuous features in the corresponding feature map.
As a result, they are usually miss-detected or detected with low confidence scores, e.g., less than the pre-defined detection threshold. 
As the $maxout$ function does not increase the values in the to-be-fused feature maps, these features will remain inconspicuous in the fused feature map.
Therefore, these objects will still be miss-detected or detected with low confidence scores.
This issue needs to be addressed; otherwise, one major benefit of cooperative perception, namely extending the effective detection range of individual autonomous vehicles, will be compromised. 

%Here, we assume objects are only partially occluded, i.e., limited amount of points of cloud can be collected from them, and feature extractor is able to extract feature from them. 

%When AVs are requiring information from other AVs, the accurate detection on these far/occluded objects are more meaningful. However, even senders is closer than receiver on some objects, it is not guaranteed to have strong enough feature for fusion results and further accurate detection.It is understandable that the sender may not have a clear sensing as well even closer to objects, and in most cases receiver do not have another better sender. F-Cooper did not consider the enhancement for weak feature, and limits their detection precision on far objects.

%%% ADD TO FEATURE BASED WEIHGT !!!!!!!!!!! We further explore why this happens when distance increase. In practice, if the two AVs are far from each other, then the correlation between two feature maps become smaller. Correspondingly, Receiver should more relays on the feature map they receive since they may not have clear look on far objects. For same region, since sender usually have clearer look, senders' feature map will have distinct feature, and receivers' have relatively weak feature. However, \textit{maxout} only output the overlap area with greater value on feature, and weak feature from receiver may effect the fusion results. 

\vspace{-15pt}
\section{CoFF: Cooperative Spatial Feature Fusion for 3D Object Detection}
To address the above-mentioned limitations, we propose the Cooperative Spatial Feature Fusion (CoFF) for cooperative 3D object detection on CAVs.
CoFF effectively integrates feature maps so that the distinctive features are kept and enhanced, while noise features are suppressed.
In essence, CoFF enables a vehicle (referred to as the receiver) to effectively utilize the supplementary information provided by another vehicle (referred to as the sender), and weighs the sender's feature map in the regions where its own feature map has a hard time detecting objects. 
With the increased weight from the sender's feature map, the noisy features on the receiver's feature map are eliminated by the \textit{maxout} function, thus improving the object detection performance.
Moreover, for objects that are either occluded or far from the receiver, CoFF enhances their corresponding features in the fused feature map, thus improving detection on these objects as well.
%To address this issue, it is crucial to enhanced the distinctive features after fusion, which improve the detection performance on far/occluded objects.
%
The CoFF approach can be represented in the following equation:
\begin{equation}
    \mathbf{F} = \left\{ \mathbf{F}_3 \cup max \{ \mathbf{F}_1, \mathbf{F}_2 \times X\} \right\} \times Y,
\end{equation}
where $\mathbf{F}_1$ and $\mathbf{F}_2$ are the overlapping areas of the two to-be-fused feature maps from the receiver and the sender, respectively, $\mathbf{F}_3$ is the non-overlapping area of the receiver's feature map, and $\mathbf{F}$ is the resulting fused feature map. $X$ is the assigned weight to the sender's feature map $\mathbf{F}_2$, and $Y$ is the feature enhancement parameter.
\vspace{-11pt}
\subsection{Information-based Spatial Feature Fusion}

%We further explore why the detection performance decay rapidly when distance increase. In practice, if the two AVs are far from each other, then the correlation between two feature maps become smaller. Correspondingly, receiver should more rely on the feature map they receive since they may not have clear look on far objects. For the same region, since sender usually have clearer look, senders' feature map will have distinct feature, and receivers' have relatively weak feature. However, \textit{maxout} only output the overlap area with greater value on feature, and weak feature from receiver may effect the fusion results. 

\begin{comment}
\begin{figure*}[!htbp]
\centering
\vspace{-11pt}{%
\includegraphics[width=55mm,height=55mm]{images/sparseconv.png}
}
\vfill
\caption{sparse feature extractor}
\vspace{-5pt}
\label{fig:sfx}
\end{figure*}
\end{comment}

As the distance between a sender and a receiver increases, the overall object detection performance of F-Cooper decreases rapidly. 
This is due to the fact that F-Cooper treats all feature maps with the same weight regardless of their information contribution. For that reason, useless features contained in the receiver's feature map is also included in the fused feature map, which leads to a decrease in object detection performance. 
In this section, we propose the information-based spatial feature fusion to address this issue.
Information-based feature fusion consists of the following two steps. 
(1) When a receiver vehicle receives a feature map shared from a sender vehicle, it first measures all the new features that are contained in the feature map, by comparing them with the features in its own feature map. 
(2) Based on the measurement, the receiver applies a weight, $X$, to the received feature map before fusion. Therefore, the fusion counteracts the negative effect caused by the weak features in the receiver's feature map.
%As we mentioned previously, the weak features in the receiver's feature map become noise after the fusion, so they affect the final detection performance.
%On the other hand, the features from the sender's feature map are more meaningful to cooperative perception, especially when the receiver's sensors are blocked. 
%
To reduce noise in the receiver's feature map, we increase the weight of the sender's feature map in proportion to how much new semantic information it can contribute to the receiver. 
The information-based fusion strategy can be expressed as 
\begin{equation}
    \mathbf{F}^{i}=max\left \{ \mathbf{F}_1^i, \mathbf{F}_2^i \times X \right \}, \forall i=1, 2, \cdots ,128 , 
\end{equation}
where $i$ denotes the index of a channel in a feature map, $\mathbf{F}_1$ represents the overlapping area of the feature map from the receiver, $\mathbf{F}_2$ represents the overlapping area of the feature map from the sender, and $\mathbf{F}^i$ is the $i$-th channel of the fused feature map. 

%Cooperative perception are desperately needed when AVs' do not have a clear sensing range due to the occlusion or distance. 
%While the receiver's feature map may only contains very limited semantic information for detection, achieving cooperative perception relies on how much new information can be provided from received feature maps. Benefiting from the location information of 3D data, the overlapping area of the two feature map can be quickly found. 

\begin{figure*}[!htbp]
\centering
\vspace{-11pt}
\subfloat[Vehicle 1 (Receiver)]{%
\includegraphics[width=44mm,height=60mm]{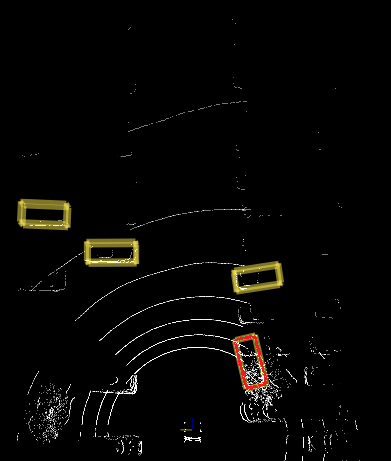}
}
\subfloat[Vehicle 2 (Sender)]{%
\includegraphics[width=44mm,height=60mm]{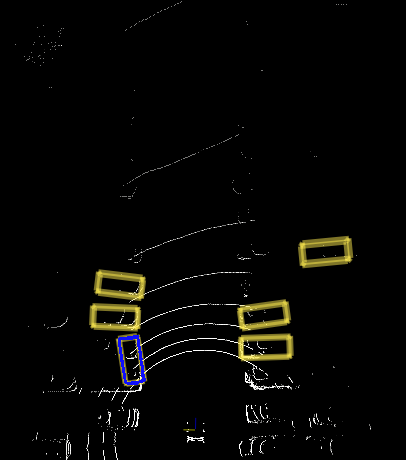}
}
\subfloat[F-Cooper detection result]{%
\includegraphics[width=44mm,height=60mm]{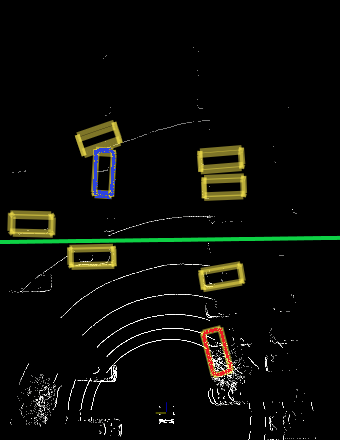}
}
\subfloat[CoFF detection result]{%
\includegraphics[width=44mm,height=60mm]{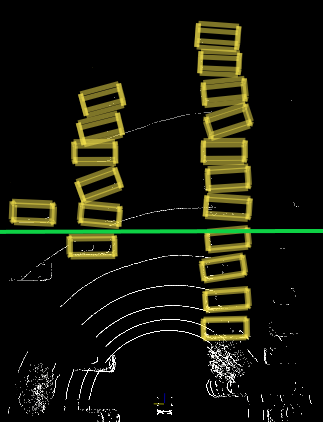}
}
\vfill
\caption{False detection correction over individual detection results. Blue and red boxes show two false detections on F-Cooper detection results, where red box is not in the fusion area. Both can be corrected by CoFF. The green line splits the fusion (above green line) and non-fusion (below green line) areas in the fused feature map.}
\vspace{-10pt}
\label{fig:correction_1}
\end{figure*}

%There is usually an overlapping region on the to-be-fused feature maps, so 
Features within the overlapping area between the sender's and receiver's feature maps are similar to each other.
%We know the sender has a clearer view of this region, as shown in Fig.~\ref{fig:limit_f}, so more semantic information is contained in the sender's feature map.
%
Therefore, this similarity between the feature maps can be used to quantify the volume of new semantic information provided by the sender.
The larger the similarity, the less supplementary information is provided by the received feature map.
We use L2 distance (also called Euclidean distance) between the corresponding features in the overlapping area of two feature maps to represent their similarity.
A large L2 distance implies that the sender's feature map is able to provide a large volume of new semantic information. On the other hand, a small L2 distance suggests that the receiver's feature map is similar to the sender's, thus new information contributed by the sender would be limited.
Besides similarity, the weight factor $X$ is also affected by the size of the overlapping area.
A larger overlapping area results in a smaller weight.
This is because a large overlapping area suggests a closer physical distance between the sender and the receiver, thus less new semantic information can be provided by the received feature map. 
%Therefore, the average L2 distance is more suited to represent the similarity than the overall distance between feature maps. 
%
The weight factor $X$ is calculated by the following formula 
\begin{flalign}
X_{}=\left\{
\begin{array}{rcl}
& S/(A_{o}/A) + 1.2 & {S < 0.15},\\
& S/(A_{o}/A) + 1.5 & {0.15 \leq S < 0.3} ,\\
& 1.8 & {S \geq 0.3 }\\
\end{array} \right.
\end{flalign}
\begin{comment}
\begin{equation}
 {S = \norm{\mathbf{F}_{1}^{i}-\mathbf{F}_{2}^{i}}/(W\times H)} ,
\end{equation}
\end{comment}
%

where ${S = \norm{\mathbf{F}_{1}^{i}-\mathbf{F}_{2}^{i}}/(W\times H)}$ and $\norm{\mathbf{F}_{1}^{i}-\mathbf{F}_{2}^{i}}$ is the L2 (or Euclidean) distance between two feature maps which are represented by two vectors $\mathbf{F}_{1}^{i}$ and $\mathbf{F}_{2}^{i}$. $W$ and $H$ are the width and height of the overlapping area of  two feature maps. \textit{A$_{o}$} is the size of the overlapping area, and \textit{A} is the size of entire feature map. The constant numbers in the above equations, e.g., $0.15$, $0.3$ and $1.2$, are derived from intensive experiments on our autonomous platform. %These numbers may vary should information-based be applied on other datasets, e.g., the KITTI dataset~\cite{Geiger2013IJRR}. 

\vspace{-11pt}
\subsection{Feature Enhancement}

Objects with weak features in the fused feature map remain hard to detect with state-of-the-art 3D object detection models. 
Inspired by the recent work proposed in \cite{wang2019side} where a binary classifier is used to predict the boundaries of objects, we discover that distant/occluded objects can be detected by increasing the difference between the values in a feature map that correspond to objects and the background.
To this end, we propose the feature enhancement mechanism which can be represented as 
\begin{equation}
 \label{eq:enhance}
 \mathbf{F} = \{ \mathbf{F}^i \times Y \}, \forall i=1, 2, \cdots ,128 , 
\end{equation}
where $\mathbf{F}\in \mathbb{R}^{C\times H \times W}$ is the enhanced feature map, $\mathbf{F}^i$ represents the $i$-th channel of the 3-dimensional feature map produced by our information-based feature fusion mechanism. 
Here, the values of $W$, $H$ and $C$ denote the width, height and the total number of channels of the fused feature map.
With Eq.~\ref{eq:enhance}, the fused feature map is enhanced by factor $Y$ before being passed to the region proposal network (RPN)~\cite{ren2015faster}, which is in charge of detecting objects from the fused feature map.
The enhancement increases the values in the feature map that represent objects, including distant and/or occluded ones.
%emerge out of background along the X, Y, and Z axles.
%
The values corresponding to the background in the feature map are all zero; as a result, Eq.~\ref{eq:enhance} does not enhance background features.

After carefully analyzing large number of feature maps, we find that the values in a strong feature are usually double or triple times larger than those in a weak feature.
%in a strong feature map are at least double or triple times larger than those in a weak feature map.
%
This observation guides us deciding the suitable values for the enhancement parameter, $Y$. 
We find that an enhancement parameter $Y=2$ or $Y=3$ is adequate for enhancing the feature maps generated from the point cloud data collected by a 16-beam LiDAR.
A larger value of $Y$ (e.g., $Y>5$) may excessively enhance the fused feature map, resulting more false detection.
\begin{figure*}[!htbp]
\centering
\vspace{-11pt}
\subfloat[Vehicle 1 (Receiver)]{%
\includegraphics[width=44mm,height=60mm]{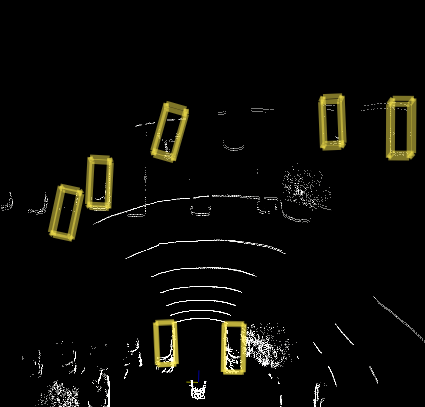}
}
\subfloat[Vehicle 2 (Sender)]{%
\includegraphics[width=44mm,height=60mm]{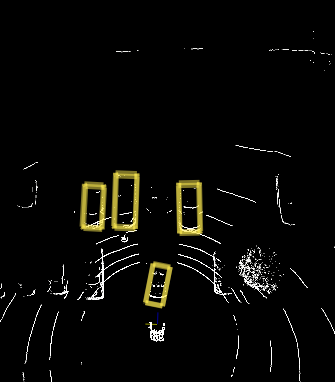}
}
\subfloat[F-Cooper detection result]{%
\includegraphics[width=44mm,height=60mm]{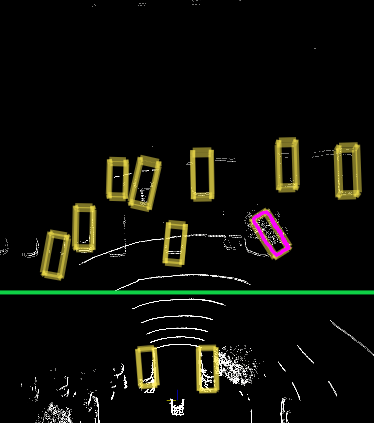}
}
\subfloat[CoFF detection result]{%
\includegraphics[width=44mm,height=60mm]{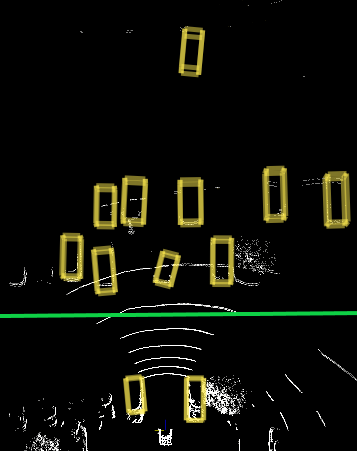}
}
\vfill
\caption{False detection correction over F-Cooper detection results. The magenta box is a false detection caused by F-Cooper's fusion function, which is corrected after the feature enhancement of CoFF. The green line splits the fusion (above green line) and non-fusion (below green line) areas in the fused feature map.}
\vspace{-10pt}
\label{fig:correction_2}
\end{figure*}
\vspace{-11pt}
\subsection{Benefits of CoFF on Object Detection}
In this subsection, we describe the benefits of applying CoFF in detecting 3D objects on autonomous vehicles.

\subsubsection{Extension of Detection Range}
One major benefit of applying CoFF to object detection is the extension of individual vehicle's detection range. 
As shown in Fig.~\ref{fig:correction_1}(a) through (c), more objects in a larger area are detected after the sender's and receiver's feature maps are fused by F-Cooper. 
%Most of points sent from receiver are blocked by near objects, and the detection on far objects relies on how much new information provided by received feature map. 
%As we can see on Fig.~\ref{fig:correction_1} (c), 
%
The detection range on the receiver is already extended by incorporating the feature map shared from the sender. 
However, the receiver struggles with detecting objects that reside at the boundary of the sender’s sensing range, as exemplified by the miss-detected object shown in blue box in Fig.~\ref{fig:correction_1}(c).
Fig.~\ref{fig:correction_1}(d) shows that after feature enhancement, CoFF is able to detect objects that are far away from both the sender and the receiver.
%The overall detection range is enlarged greatly by the feature enhancement. 
%
As those features on the receiver's feature map are relatively weak, such improvement is mostly contributed by the enhanced features in the sender's feature map.
The extension of the detection range is more significant in scenarios where less point cloud data is collected on an autonomous vehicle, e.g., due to occlusions or low-resolution data collected by a low-end LiDAR sensor.

\subsubsection{Enhancement on Detection}
After enhancing the fused feature map, false detection originally caused by weak features can be reduced to a certain extent.
%As a result, CoFF is able to correct certain false detection results existed in the original CNN-based detection models.
%False detection issue appear frequently in 3D detection, and most of them are caused by the indistinct representation of feature. As our feature enhancement approach well enhance the weak feature from feature map, false detection issue can be fixed after the enhancement. In general, false detection may occur on the results of individual detection, or the fusion results of F-Cooper. Both of them can be corrected by our feature enhancement approach. 
%
A false detection example is shown in Fig.~\ref{fig:correction_1}(b) where the blue box indicates that one vehicle is detected in this area. 
However, in reality there are two vehicles located in the blue box.
Zooming into the blue box area, we identified the front portions of two vehicles that are parked parallel to each other. 
When the LiDAR sensor scanned this area, most of the laser signals are blocked by the vehicle(s) in front of them, resulting in little or no point cloud data collected from the rear portions of the two vehicles.
Since the two vehicles are close to each other, the point cloud data collected from the front portions of the vehicles are treated as a whole, i.e., a single vehicle was mistakenly detected.

%This type of occlusion is not rare in a real-world setting, therefore, such false detection may occur frequently.
%
This issue can be addressed by CoFF so that the two vehicles can be separated, as shown on Fig.~\ref{fig:correction_1}(d).
This is because the weak features representing the rear portions of these two vehicles are enhanced by CoFF.
With a stronger feature representing each vehicle, the RPN is able to detect two vehicles, instead of mistakenly treating the weak features as background.
%The correction from feature enhancement is not an occasional case in our experiment. 
%
Moreover, the red box marked in Fig.~\ref{fig:correction_1}(a) is another false detection case which originally is an overhanging tree and is not in the fusion region. Such false detection cases can also be reduced after feature enhancement, indicating that the enhancement is effective on individual detection models as well. 

%%%%%%%%%%%%%%%%%%%%%%%%%%%%%%%%Based on the analysis above, we believe our feature enhancement enhance the feature of the back part of those two vehicles, and instead of ignoring that part as background, detectors are now able to accurately detect these vehicles. 
%In experience, the range of enhancement parameter is from 2 to 6. A higher value than 6 may cause false detection, specifically on models with the capacity to detect various objects such as vehicles, pedestrian, street lamp or traffic sign.

\subsubsection{Enhancement on Fusion}

Another type of false detection may arise when two weak features are improperly fused by the $maxout$ function.
Because the $maxout$ function selects the most prominent features from the to-be-fused ones, some weak features may be enhanced in a wrong way.  
This is a unique problem to F-Cooper, because two feature maps are directly fused by the $maxout$ function.
%This is a fairly common problem when two feature maps are not carefully fused.
%
Fig.~\ref{fig:correction_2}(c) shows an example of such false detection, in which a non-existing vehicle (depicted in the magenta box) appears after the feature maps generated by the sender and the receiver are fused by the $maxout$ function. 
Within the magenta box, there is a tree (the ground truth) which was falsely detected as a vehicle.
With the proposed CoFF, the resulting feature map better represents the object in this area, as shown in Fig.~\ref{fig:correction_2}(d), thus avoiding the false object detection. 

{
\color{red}
\begin{table*}[!thbp]
\centering
%{|p{5mm}|p{5mm}|p{5mm}|p{5mm}|p{5mm}|p{5mm}|p{5mm}|p{5mm}|p{5mm}|p{5mm}|p{5mm}|}
%\begin{tabular}{|p{25mm}|p{10mm}|p{6mm}|p{6mm}|p{6mm}|p{6mm}|p{6mm}|p{6mm}|c{6mm}|c{6mm}|p{6mm}|p{6mm}|p{7mm}|p{7mm}|}
\begin{tabular}{|c|c|c|c|c|c|c|c|c|c|c|c|}
\hline
Scenario&Dataset&\multicolumn{2}{c|}{Cooper~\cite{chen2019cooper}}&\multicolumn{2}{c|}{F-Cooper~\cite{chen2019f}}&\multicolumn{2}{c|}{CoFF w/o Enhancement}&\multicolumn{2}{c|}{CoFF}&\multicolumn{2}{c|}{Improvement}\\
\cline{3-12}
&&Near&Far&Near&Far&Near&Far&Near&Far&Near&Far\\
\hline
\hline
Multi-lane Roads&KITTI&77.46&71.42&65.51&52.78&~~~75.86~~~~&56.76&82.75&70.27&26.32&33.14\\
\hline
Road Intersections&T\&J &80.21&72.37&54.28&28.57&~~~62.75~~~~&34.61&72.54&59.25&33.64&107.38\\
\hline
Parking Lots&T\&J&71.88&56.94&51.85&21.05&~~~60.71~~~~&28.57&64.28&57.14&23.97&171.44\\
\hline
\end{tabular}
\caption{Precision comparison among Cooper, F-Cooper, CoFF, and the improvement of CoFF over F-Cooper on receiver (\%).}
\vspace{-11pt}
\label{Table:compare}
\end{table*}
}

\vspace{-11pt}
\section{experiment and results evaluation}

In order to compare performance with F-Cooper under the same settings, we carry out experiments with the T\&J dataset used in~\cite{chen2019f} for cooperative 3D object detection. Due to the limited cases available in the original T\&J dataset, F-Cooper only provides performance evaluation in limited scenarios. During this work, we enrich the T\&J dataset with the following three scenarios: road intersections, multi-lane roads and parking lots. Road intersection and multi-lane road scenarios are used to simulate real driving conditions on road, and parking lot scenarios are used to evaluate CoFF in more cases with crowded environments. The T\&J dataset is collected using a 16 beams LiDAR. To evaluate CoFF on high-resolution 3D data, we select appropriate cases from the KITTI dataset~\cite{Geiger2013IJRR}, a well-known dataset for 3D objects detection, to make a general comparison between CoFF and F-Cooper. In the evaluation, we define all objects within 20 meters from the receiver AV to be in the ``near'' category, and the objects beyond this range to be in the ``far'' category.

The evaluated dataset contains more than 1,500 and 200 sets of data from the T\&J and KITTI datasets, respectively. In experiment, we compare our CoFF solution with the feature fusion approach from F-Cooper, and the spatial feature map corresponds to a 3D space with a range of $[0,70.4]$, $[-40,40]$ and $[-3,1]$ meters along the $x$, $y$, and $z$ axles. Any point clouds exceeding this range will have no features on the feature map. Our equipment to run CoFF is a desktop equipped with a NVIDIA Quadro P4000 GPU.

\vspace{-11pt}
\subsection{Improvement from Fusion}

CoFF consists of two parts, information-based fusion and feature enhancement. 
Information-based fusion is an 
improvement to the original \textit{maxout} fusion used in F-Cooper, while feature enhancement is our novel approach to 3D object detection. We want to first quantify the improvement made possible from only information-based fusion. By doing so, we are able to find an upper-bound limit for the current 3D feature fusion strategy. Therefore, we discuss the improvements in detection precision without feature enhancement in this subsection. 

Table~\ref{Table:compare} shows the detection precision of CoFF without feature enhancement, where both Intersection over Union (IoU) and confidence score threshold is set at 0.5. For the ``near'' category, F-Cooper achieves a relatively high precision in open area scenarios such as multi-lane roads and road intersections.
Due to the high similarity between the feature maps generated from two nearby vehicles, they are treated equally by \textit{maxout}, i.e., the resulting fused feature map does not effect much on the detection results.
Even so, our information-based fusion is able to obtain improvement of approximately 10\% for multi-lane roads and 8\% for road intersections. For parking lot scenarios, due to occlusion, both feature maps (from sender and receiver) contain less distinct features than that of open area scenarios. Thus, we see a decrease in detection precision for F-Cooper. For CoFF, the weights of the sender's feature map on fusion in these cases are much higher than those in an open area scenario, due to the low similarity between the two feature maps. Therefore, with more semantic information supplied by the received feature maps, CoFF is not affected much by occlusion, and still able to achieve approximately 9\% improvement on precision. 
%CoFF is showing more improvement in the parking lot scenarios. 

For the ``far'' category, as distance increases, the similarity between the two to-be-fused feature maps decreases accordingly. We see a lower detection precision with F-Cooper in all cases, especially in the parking lot scenarios. The improvements of our method over F-Cooper are slightly lower than those for the ``near'' category, which is about 7\% for cases from T\&J dataset and 4\% for cases from KITTI dataset. 
%Closer objects from the sender will have distinct features on the feature map. They also have a high probability to keep said features after the information-based fusion. 
%Objects closer to the sender will have more distinct features on the sender's feature map. Such features also have a higher probability to be retained in information-based fusion. 
Objects closer to the sender will have more distinct features on the sender's feature map. Such features also have a higher probability to be retained in information-based fusion. 
In the experiment, we found that most newly detected objects by the receiver is closer to the sender, which means our information-based fusion works well in retaining important features from the sender's feature map, thus improving cooperative perception. 
However, for objects that are far from both the sender and the receiver, their features are weak on both original feature maps and thus stay weak after fusion, therefore most of them remain undetected. 
\vspace{-11pt}
\subsection{Scenario Evaluation}

To make a fair comparison of our CoFF method and how much improvement it yields, we re-implemented F-Cooper as our baseline. Moreover, we utilize Cooper fusion on the original 3D data as a reference for upper-bound limit in the evaluation. 
In the experiment, we follow the same approach as F-Cooper in designing our evaluation scenarios but with more evaluation cases. We report the precision by comparing the true detected vehicles against the ground truth, and set both the Intersection over Union (IoU) and confidence score threshold at 0.5 for detection. We also provide the corresponding improvement percentages over F-Cooper for a clear comparison.

%and for each scenario, we have more  set the Intersection over Union (IoU) threshold at 0.5 for detection, and also provide the corresponding percentage of improvement compared with F-Cooper.
%and also provide the results at IoU=0.3 to show the full-scale detection ability of CoFF. 
%All scenarios are strictly follow F-Cooper with more evaluate cases. Precision are reported by comparing true detected vehicles with ground truth. 

In Table \ref{Table:compare}, we first take a look at the baseline, which is F-Cooper. F-Cooper makes use of a non-weighted feature fusion which achieves a relatively high detection precision in the ``near'' category. We observe that F-Cooper is able to perform well on both multi-lane roads and road intersections, which shows that F-Cooper works well in open area scenarios and in short distance cases; the precision in these scenarios are over 54\%. However, due to occlusions, the precision of F-Cooper for parking lots is not as good as compared to other scenarios. We see a precision drop in the parking lots cases with the precision being 51.85\%. By comparison, the effect of occlusion is not as obvious on the CoFF method, CoFF is able to achieve a precision of 64.28\% in the parking lot cases. For open area cases, the precision is above 72\% on both multi-lane roads and road intersections, which shows a great improvement. %the effect of occlusion is not obvious on the CoFF method, which is able to achieve improvement in all scenarios, especially in the parking lot cases. 
The precision of CoFF in all scenarios is close to the upper-bound, Cooper, which means we could achieve a similar detection performance as raw sensor data fusion with much less data being transmitted between vehicles. Note that the precision of CoFF for multi-lane road scenario exceeds that of Cooper. This is attributed to the improvement made from feature enhancement of CoFF, while Cooper shows the upper-bound results without feature enhancement.

Moving to the ``far'' category, as distance increases, the precision of F-Cooper decreases rapidly due to the noisy feature's effect. In the road intersection cases, the precision of F-Cooper is 28.57\%. In the multi-lane road cases, since data from KITTI was collected by a high-end LiDAR, F-Cooper is able to achieve a precision of 52.78\%. For more crowded scenarios such as the parking lot cases, the precision of F-Cooper drops to 21.05\%. In contrast, the CoFF results clearly demonstrate the benefit of its feature enhancement improvement over F-Cooper, and even exceeding Cooper (raw sensor data fusion) in some cases. The precision of CoFF is above 60\% for both scenarios from T\&J dataset. Furthermore, CoFF does not present an obvious precision decline in parking lot cases. Benefiting from feature enhancement, most occluded objects are detected with a higher confidence score, suggesting that CoFF is not as sensitive to occlusion compared to F-Cooper. 

In real-word deployment scenarios, a CAV participating in cooperative perception most likely would receive incoming sensor data from multiple, instead of one, close-by vehicles. For that reason, it is necessary to further evaluate the effectiveness of CoFF in multiple senders scenarios. For cases with two senders, CoFF preforms slight better in the ``near'' category, with approximately 18\% higher precision improvement compared with the same cases with one sender. When more senders participate in cooperative perception, more semantic information can be obtained and fused from a larger physical perception region. We thus achieve a more prominent improvement on detecting ``far'' objects, with about 40\% more improvement on precision. The improvement is greater in scenarios where there are more occlusions and CoFF is able to see a larger detection range in multiple senders cases. Due to space limitation, we here omit the detailed evaluation of CoFF for multiple senders cases.

\vspace{-11pt}
\subsection{Precision Evaluation}

Benefiting from the enriched T\&J dataset, we are able to provide an overall accurate evaluation of CoFF's precision under various scenarios. Even though F-Cooper performs well on ``near'' objects, we are still able to see improvements with CoFF. \textbf{For CoFF, 70.51\% of vehicles in the ``near'' category can be correctly detected, while F-Cooper achieves 52.13\% within the same category.} 
For the ``far" cases, CoFF has a even greater improvement over F-Cooper. \textbf{F-Cooper only has an average of 23.07\% of detection precision in the ``far'' category, with most of the successful detection cases being within 35 meters. In contrast, CoFF is able to achieve a 58.18\% detection precision, and the effective detection range reaches up to 50 meters.} This improvement is mainly from our feature enhancement approach. It is worth mentioning that the T\&J dataset is collected by a 16-beam LiDAR, which generates fairly sparse point cloud data for far distance objects. Our experiment results suggest that CoFF's improvement works well in complicated road scenarios with limited points received by the LiDAR, and does not require dense 3D point cloud data.
%The improvement may not be this obvious if apply on a more dense dataset, e.g., KITTI. 

\begin{comment}
\begin{table}[h]  % [h]表格在文中放置的位置
  \centering  %作用是使表格居中
  \begin{spacing}{1.35}  %调整表格行距
  \caption{Overall Precision Comparison} %表格标题
  \label{general} %表格标签 方便引用
  \resizebox{0.7\hsize}{!}{  % 缩小整体表格 需要导包
    \begin{tabular}{|c|c|c|}   % c 表示表格中的文字居中
    \hline
   \diagbox{Area}{Approach} & F-Cooper & CoFF \\  
    \hline
      Near & 54.43 & 77.37  \\
    \hline
      Far &  23.71 & 59.74 \\
    \hline
    \end{tabular}
  }
  \end{spacing}
\end{table}
\end{comment}

\begin{figure}[!htbp]
    \centering
    \includegraphics[width=0.9\linewidth]{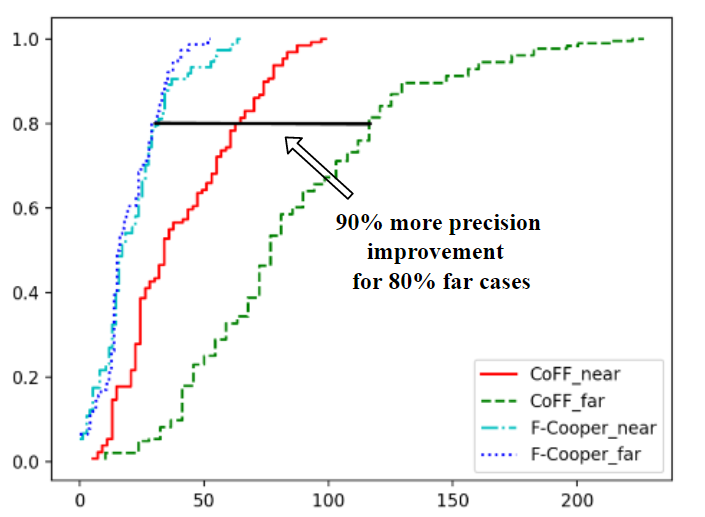}
    \vspace{-3pt}
    \caption{Cumulative Distribution Function vs. detection precision improvement percentages over individual vehicles}
    \vspace{-10pt}
    \label{fig:cdf}
\end{figure}

To clearly illustrate how much improvements achievable with CoFF, we compare CoFF with F-Cooper in the ``near'' and ``far'' categories, respectively. Fig.~\ref{fig:cdf} shows the Cumulative Distribution Function (CDF) of the precision improvement over the detection results of individual vehicles.
As shown in the figure, for the ``near'' category, CoFF is able to achieve a 65\% precision improvement for 80\% of cases, while F-Cooper has about 32\% of improvement when comparing with the detection results on individual vehicles. 
When it comes to the ``far'' category, CoFF is able to achieve a more distinct improvement. The improvement achieved by F-Cooper is around 20\% for over 60\% of cases, and is within 40\% for almost 90\% of cases. By comparison, CoFF is able to achieve about 120\% of detection improvement over 80\% of cases.

\vspace{-11pt}
\subsection{Detection Range Evaluation}

\begin{figure}[!htbp]
    \centering
    \includegraphics[width=0.9\linewidth]{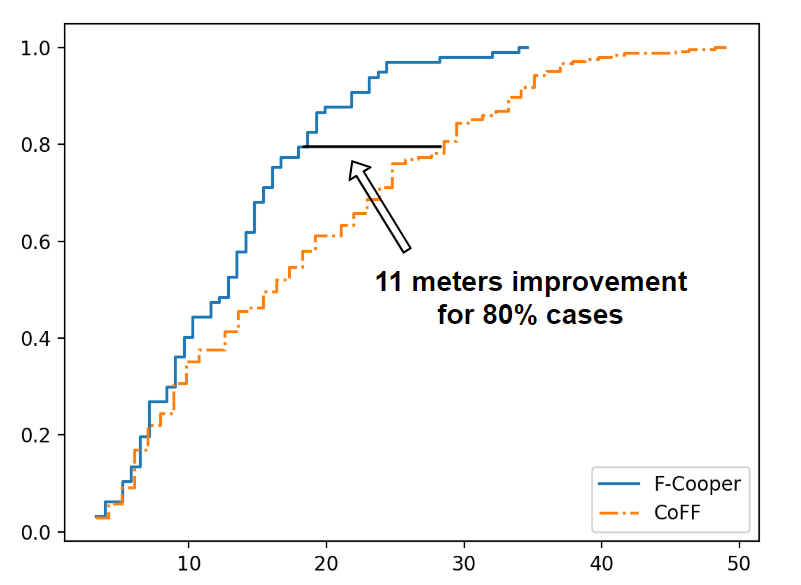}
    \vspace{-7pt}
    \caption{Cumulative Distribution Function vs. range of detected objects in meters}
    \vspace{-15pt}
    \label{fig:cdf_range}
\end{figure}

Through feature enhancement, CoFF is able to enhance weak features for far objects. Therefore, the effective detection range of CoFF is extended greatly over F-Cooper. We illustrate this improvement in Fig.~\ref{fig:cdf_range} which shows drastic difference in detection range of the two approaches. For objects detected by F-Cooper, 83\% of them are within 20 meters, which is in the ``near'' category. In contrast, for CoFF, the objects belonging to the ``near'' category only represent 61\% of all detected. As we look deeper, the maximum detection range of F-Cooper is around 35 meters, while CoFF is up to 50 meters. 
%In addition, 39\% of detected objects by CoFF are in the ``far'' category, while F-Cooper is 17\% in the same category. 
For 80\% cases evaluated by the two approaches, CoFF achieves an average of 11 meters of detection range improvement over F-cooper. As high resolution 3D data typically contains more dense points for far objects, this improvement is prominent in cases with low-resolution 3D data.

\vspace{-11pt}
\subsection{Detection Threshold Evaluation}

\begin{figure}[!htbp]
    \centering
    \includegraphics[width=0.9\linewidth]{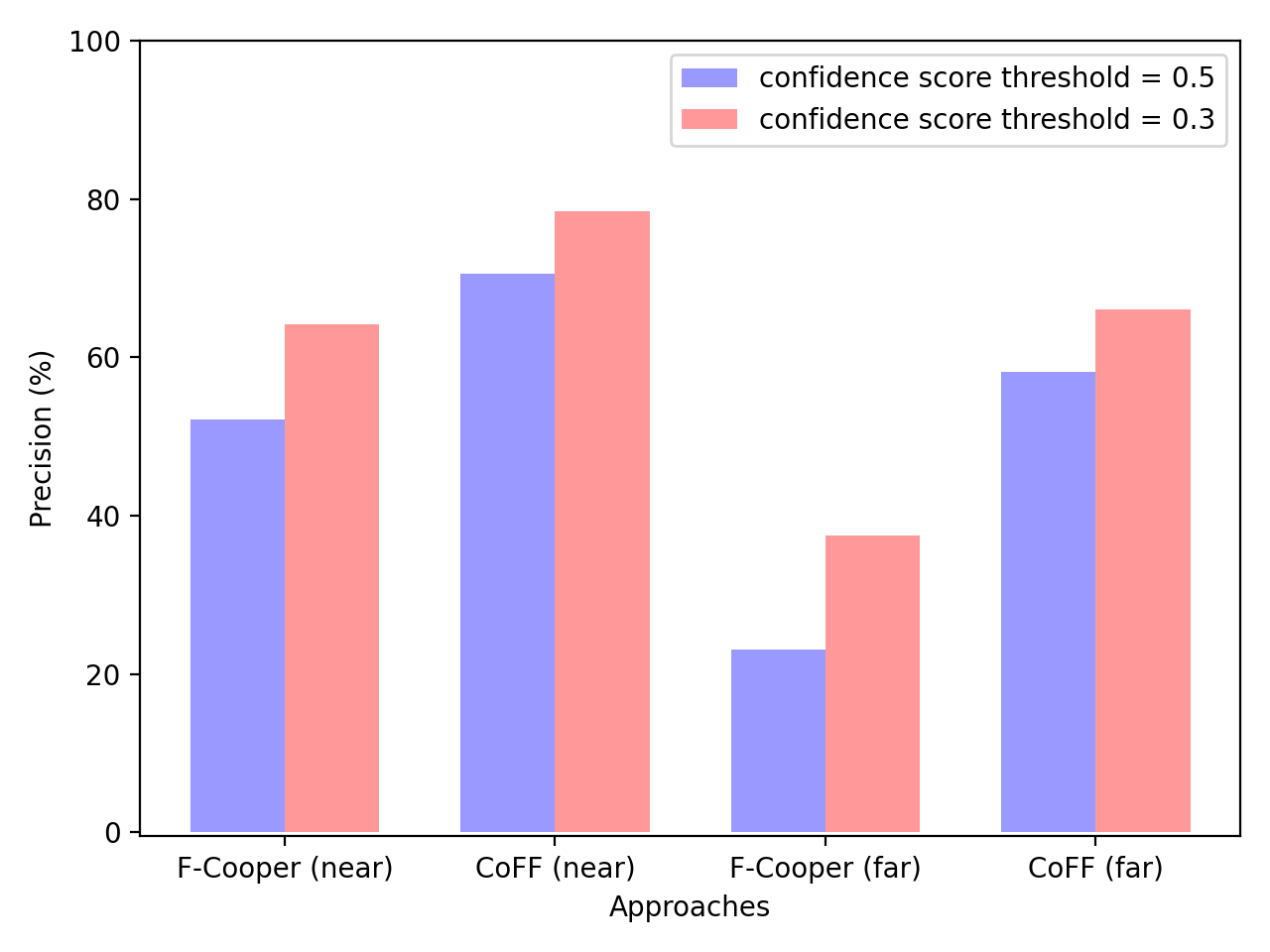}
    \vspace{3pt}
    \caption{Precision comparison between F-Cooper and CoFF over different confidence score thresholds}
    \vspace{-12pt}
    \label{fig:thresholds}
\end{figure}

\begin{comment}
\begin{table*}[h]  % [h]表格在文中放置的位置
  \centering  %作用是使表格居中
  \caption{Precision Comparison over Detection Thresholds} %表格标题
  \label{threshold} %表格标签 方便引用
  \resizebox{0.7\hsize}{!}{  % 缩小整体表格 需要导包
    \begin{tabular}{|c|c|c|c|c|}   % c 表示表格中的文字居中
    \hline
   \diagbox{Area}{Threshold} & F-Cooper (0.5) & CoFF (0.5) & F-Cooper (0.3) & CoFF (0.3)\\  
    \hline
      Near &54.43& 75.37 & 69.22&81.55 \\
    \hline
      Far &23.71& 59.74 & 35.51&67.19 \\
    \hline
    \end{tabular}
  }
  
\end{table*}

\begin{figure*}[!htbp]
\centering
\vspace{-11pt}
\subfloat[Groundtruth]{%
\includegraphics[width=36mm,height=60mm]{images/gdth.png}
}
\subfloat[Car 1 (Receiver)]{%
\includegraphics[width=36mm,height=60mm]{images/60_paper.png}
}
\subfloat[Car 2 (Sender 1)]{%
\includegraphics[width=36mm,height=60mm]{images/121_paper.png}
}
\subfloat[Car 3 (Sender 2)]{%
\includegraphics[width=36mm,height=60mm]{images/177_paper.png}
}
\subfloat[CoFF]{%
\includegraphics[width=36mm,height=60mm]{images/60-121_177_paper.png}
}
\vfill
\caption{Detection result of CoFF on multiple sender cases}
\vspace{-5pt}
\label{fig:fusion_3}
\end{figure*}

\end{comment}

Confidence score plays a great role in determining the making of a great detection model, as it directly impacts the model's final performance. Different thresholds impact the performance of certain models greatly, as seen in the performance of F-Cooper with different detection thresholds. 
However, CoFF is less sensitive to the confidence threshold than F-Cooper. Fig.~\ref{fig:thresholds} shows the precision comparison between F-Cooper and CoFF, with IoU $= 0.5$. We set the thresholds as 0.3 and 0.5, respectively. Thresholds over 0.5 will cause a great drop in detecting far objects and thresholds below 0.3 introduce a large number of false detections. As shown in Fig.~\ref{fig:thresholds}, we see a great drop in precision when changing the threshold from $0.3$ to $0.5$ on F-Cooper, for both ``near'' and ``far'' categories. 
In contrast, there are minor changes on CoFF when the confidence score increases from 0.3 to 0.5, especially in the ``near'' category. We investigate the reasons in depth by analyzing the confidence scores of all detected objects by F-Cooper. As T\&J dataset is collected by a low-end 16 beam LiDAR, the largest distribution interval of confidence score (of F-Cooper) resides between 0.3 to 0.5. With feature enhancement, most of them are converted to higher confidence scores greater than 0.5. The features of most objects are enhanced well by feature enhancement, making our model less sensitive to the setting of detection threshold. 
In experiments, the confidence scores of most detected objects by CoFF are between 0.5 and 0.7. Since the improvement from a lower detection threshold is limited, we choose 0.5 as the detection threshold of CoFF to avoid false detection.

\vspace{-11pt}
\section{Conclusions}

In this paper, we propose CoFF, a novel feature map based fusion approach for achieving cooperative 3D object detection on autonomous vehicles. 
CoFF consists of two parts: information-based fusion and feature enhancement. 
While the former allocates different weights on the received feature maps according to the amount of semantic information they contribute to the fusion, the latter enlarges the difference between the object and non-object areas on the feature map to achieve a better detection performance. 
%To the best of our knowledge, CoFF is the first feature map based fusion approach for cooperative perception on AV with the tendentious enhancement on valuable feature on feature map. 
%
Experimental results show that CoFF offers a better cooperative 3D object detection performance than F-Cooper while maintaining the same advantage of reduced data transmission, and does not require high-quality of 3D point cloud data.

% or



\vspace{-11pt}
\section*{Acknowledgment}
The work is supported by National Science Foundation grants CNS-1852134, OAC-2017564, ECCS-2010332, and the research grant from Fujitsu Laboratories of America Inc.  %We also thank Connie Wang, Shang Wang, and Shreya Nakka, the TAMS (Texas Academy of Mathematics \& Science) students at University of North Texas, for their efforts on collecting and labelling training LiDAR datasets. 

\ifCLASSOPTIONcaptionsoff
  \newpage
\fi



%\bibliographystyle{IEEEtran}
% <OR> manually copy in the resultsant .bbl file

%\bibliographystyle{unsrt}
\bibliographystyle{IEEEtran}
\bibliography{ref}

\begin{IEEEbiographynophoto}{Jingda Guo}
is a Ph.D. candidate in the Department of Computer Science and Engineering at University of North Texas, Denton, TX, USA. He received his B.S. degree in Electrical Engineering from Northeast Electric Power University, China, and M.S. degree in Computer Engineering from University of Delaware, USA, in 2015 and 2017, respectively. His research interests include Internet of Things, Connected and Autonomous Vehicles, and Computer Vision.
\end{IEEEbiographynophoto}

\begin{IEEEbiographynophoto}{Dominic Carrillo}
is a Ph.D. candidate in the Department of Computer Science and Engineering at University of North Texas, Denton, TX, USA. He received his B.S. degree in Computer Science and Mathematics from Sul Ross State University, Alpine, TX, USA. His research interests include Connected and Autonomous Vehicles, and Simultaneous Localization and Mapping (SLAM).
\end{IEEEbiographynophoto}

\begin{IEEEbiographynophoto}{Sihai Tang}
is a Ph.D. candidate and a part of the Dependable Computing Systems Lab (DCS) in the Department of Computer Science and Engineering at University of North Texas, Denton, TX, USA. He received his B.S. degree in Computer Science from University of Texas at Austin in 2017, Austin, TX, USA. His research interests include Edge Computing, Federated Machine Learning, and File Systems.
\end{IEEEbiographynophoto}

\begin{IEEEbiographynophoto}{Qi Chen}
is a Ph.D. candidate in the Department of Computer Science and Engineering at University of North Texas, Denton, TX, USA. He received B.S. and M.S. degrees in Electronic Engineering from Xidian University and Northwestern Polytechnical University, China, in 2006 and 2010, respectively. His research interests include Perception on Autonomous Vehicles, Edge AI and Internet of Things.
\end{IEEEbiographynophoto}

\begin{IEEEbiographynophoto}{Qing Yang}
is an Assistant Professor in the Department of Computer Science and Engineering at University of North Texas, Denton, TX, USA. He received B.S. and M.S. degrees in Computer Science from Nankai University and Harbin Institute of Technology, China, in 2003 and 2005, respectively. He received his Ph.D. degree in Computer Science from Auburn University in 2011. His research interests include Internet of Things, Connected and Autonomous Vehicles, Network Security and Privacy. %He serves as an Associate Editor for IEEE Internet of Things, and Elsevier Vehicular Communications journals. 
\end{IEEEbiographynophoto}

\begin{IEEEbiographynophoto}{Song Fu}
is an Associate Professor in the Department of Computer Science and Engineering at University of North Texas, Denton, TX, USA. He received his Ph.D. degree in Computer Engineering from Wayne State University in 2008. His research interests include Parallel and Distributed Systems, Cloud and Edge Computing, Connected and Autonomous Vehicles, System Reliability, and Machine Learning. %He has published over 120 journal and conference articles in these areas. He is a Senior Member of IEEE, and Member of ACM.   
\end{IEEEbiographynophoto}

\begin{IEEEbiographynophoto}{Xi Wang}
is a Research Scientist with Fujitsu Network Communications in Richardson, Texas USA. He received his B.S. degree in Electronic Engineering from Doshisha University, Japan in 1998, and received M.E. and Ph.D. degrees in information and communication engineering from the University of Tokyo, Japan in 2000 and 2003, respectively. His research interests include Software-Defined Networks (SDN), Packet Optical Networks, Vehicular Networks, Optical/Wireless Network Integration, Edge Computing/AI applications, and future Information and Communications Technology (ICT) fusion.
\end{IEEEbiographynophoto}

\begin{IEEEbiographynophoto}{Nanan Wang}
is a Research Scientist at Fujitsu Network Communications in Richardson, Texas, USA. He received his B.S. degree in Computer Science and a M.E. degree in Computer Systems Organization from Jilin University, China. He also received his M.S. and Ph.D. degrees in Computer Science from The University of Texas at Dallas. His current research interests lie within Edge AI and Deep Learning for Computer Vision. His previous research experiences also includes Vehicular Networking, 5G Systems design, and Optical Network Systems. He has co-authored over 25 conference and journals papers, and holds 7 US patents.
\end{IEEEbiographynophoto}

\begin{IEEEbiographynophoto}{Paparao Palacharla}
Biography text here.
\end{IEEEbiographynophoto}

%\newpage


%\vfill


\end{document}